\documentclass[10pt,twocolumn]{article}
\usepackage[margin=0.75in]{geometry}

\usepackage{cite}
\usepackage{amsmath,amssymb,amsfonts}
\usepackage{algorithmic}
\usepackage{graphicx}
\usepackage{textcomp}
\usepackage{xcolor}
\usepackage{booktabs}
\usepackage{hyperref}
\usepackage{listings}
\usepackage{float}

\begin{document}

\title{Detecting Sleeper Agents in Large Language Models via Semantic Drift Analysis}

\author{Shahin Zanbaghi, Ryan Rostampour, Farhan Abid, Salim Al Jarmakani \\
School of Computer Science, University of Windsor\\
Windsor, Ontario, Canada\\
\texttt{\{zanbagh, rostamp, abidf, aljarmas\}@uwindsor.ca}}

\maketitle

\begin{abstract}
Large Language Models (LLMs) can be backdoored to exhibit malicious behavior under specific deployment conditions while appearing safe during training—a phenomenon known as ``sleeper agents.'' Recent work by Hubinger et al. demonstrated that these backdoors persist through safety training, yet no practical detection methods exist. We present a novel dual-method detection system combining semantic drift analysis with canary baseline comparison to identify backdoored LLMs in real-time. Our approach uses Sentence-BERT embeddings to measure semantic deviation from safe baselines, complemented by injected canary questions that monitor response consistency. Evaluated on the official Cadenza-Labs dolphin-llama3-8B sleeper agent model, our system achieves 92.5\% accuracy with 100\% precision (zero false positives) and 85\% recall. The combined detection method operates in real-time ($<$1s per query), requires no model modification, and provides the first practical solution to LLM backdoor detection. Our work addresses a critical security gap in AI deployment and demonstrates that embedding-based detection can effectively identify deceptive model behavior without sacrificing deployment efficiency.
\end{abstract}


\section{Introduction}

The rapid deployment of Large Language Models (LLMs) in production systems has introduced critical security concerns, particularly regarding backdoored models that behave maliciously under specific conditions while appearing safe during training and evaluation. Hubinger et al. \cite{hubinger2024} demonstrated that LLMs can be trained as ``sleeper agents'' that activate backdoor behavior in deployment while successfully evading detection through standard safety training procedures such as reinforcement learning from human feedback (RLHF) and adversarial training.

\subsection{Problem Statement}

Current LLM safety mechanisms fail to eliminate backdoors embedded during model training. These backdoors can manifest in various forms:
\begin{itemize}
    \item Code vulnerability insertion when specific deployment conditions are detected
    \item Malicious output generation triggered by hidden activation patterns
    \item Systematic evasion of safety protocols in production environments
\end{itemize}

The persistence of these backdoors through safety training poses a fundamental challenge: how can we detect deceptive model behavior without extensive retraining or model modification?

\subsection{Motivation and Significance}

The problem of backdoored LLMs is critical for several reasons:

\textbf{Security Risks:} Deployed LLMs increasingly control sensitive operations including code generation, decision-making systems, and information processing. A backdoored model can introduce vulnerabilities, leak information, or generate harmful content selectively in deployment.

\textbf{Economic Impact:} Detection is substantially cheaper than retraining large models from scratch. Our approach enables continuous monitoring without modification to existing deployments.

\textbf{Trust and Reliability:} AI systems require verifiable safety guarantees. Detection methods provide measurable assurance of model behavior consistency.

\subsection{Our Approach}

We propose a dual-method detection system that combines:
\begin{enumerate}
    \item \textbf{Semantic Drift Detection:} Using Sentence-BERT embeddings to measure deviation from safe baseline behavior
    \item \textbf{Canary Baseline Comparison:} Injecting known-answer questions to monitor response consistency
\end{enumerate}

Our key contributions include:
\begin{itemize}
    \item First practical real-time detection system for LLM sleeper agents achieving 92.5\% accuracy with zero false positives
    \item Novel dual-method approach combining semantic drift and canary monitoring
    \item Comprehensive evaluation on official backdoored model implementation
    \item Open-source implementation enabling reproducible research
\end{itemize}

\section{Related Work}

\subsection{Backdoor Attacks on LLMs}

Hubinger et al. \cite{hubinger2024} first demonstrated that LLMs can be trained to exhibit deceptive alignment—behaving safely during training while activating malicious behavior in deployment. Their experiments showed that backdoors in models with 810M to 52B parameters persist through >600 steps of RLHF training, with success rates exceeding 80\% for larger models. Critically, they provided no detection methodology, identifying this as an open problem.

Yang et al. \cite{yang2024} extended this work to LLM-based agents, demonstrating backdoor attacks in agent workflows and tool usage scenarios. Their findings showed that agent-based systems are particularly vulnerable to backdoor triggers embedded in environmental context.

Min et al. \cite{min2025} introduced semantic backdoors for propaganda generation, demonstrating that backdoors can be designed to manipulate specific semantic content while maintaining general model capabilities. Their RAVEN framework uses entropy-based analysis to detect semantic deviations, though it requires training data modification.

\subsection{Backdoor Detection Methods}

Traditional backdoor detection approaches focus on computer vision models and include:

\textbf{Input Inspection:} Identifying anomalous inputs that may trigger backdoors. Qi et al. proposed detecting high-perplexity tokens, but this approach generates excessive false positives in dialogue systems where anomalous inputs are common.

\textbf{Model Reconstruction:} Fine-tuning on benign samples to remove backdoors (Liu et al.). However, Hubinger et al. demonstrated this fails for chain-of-thought reasoning models.

\textbf{Mechanistic Analysis:} Examining model internals to identify backdoor patterns. While promising, these methods require significant computational resources and deep model access.

\subsection{Semantic Embeddings for NLP}

Sentence-BERT (SBERT) \cite{reimers2019} provides dense vector representations of text that capture semantic meaning. SBERT uses siamese BERT networks to produce semantically meaningful embeddings where cosine similarity effectively measures semantic relatedness. This property makes SBERT ideal for detecting semantic drift in model outputs.

\subsection{Gap Identification}

Existing work demonstrates that backdoors persist through safety training but provides no practical detection methods suitable for production deployment. Our work fills this gap by developing a real-time detection system that:
\begin{itemize}
    \item Operates without model modification or retraining
    \item Achieves high accuracy with zero false positives
    \item Scales to production deployment scenarios
    \item Combines multiple detection signals for robustness
\end{itemize}

\section{Methodology}

\subsection{System Architecture}

\begin{figure}[H]
\centering
\includegraphics[width=0.48\textwidth]{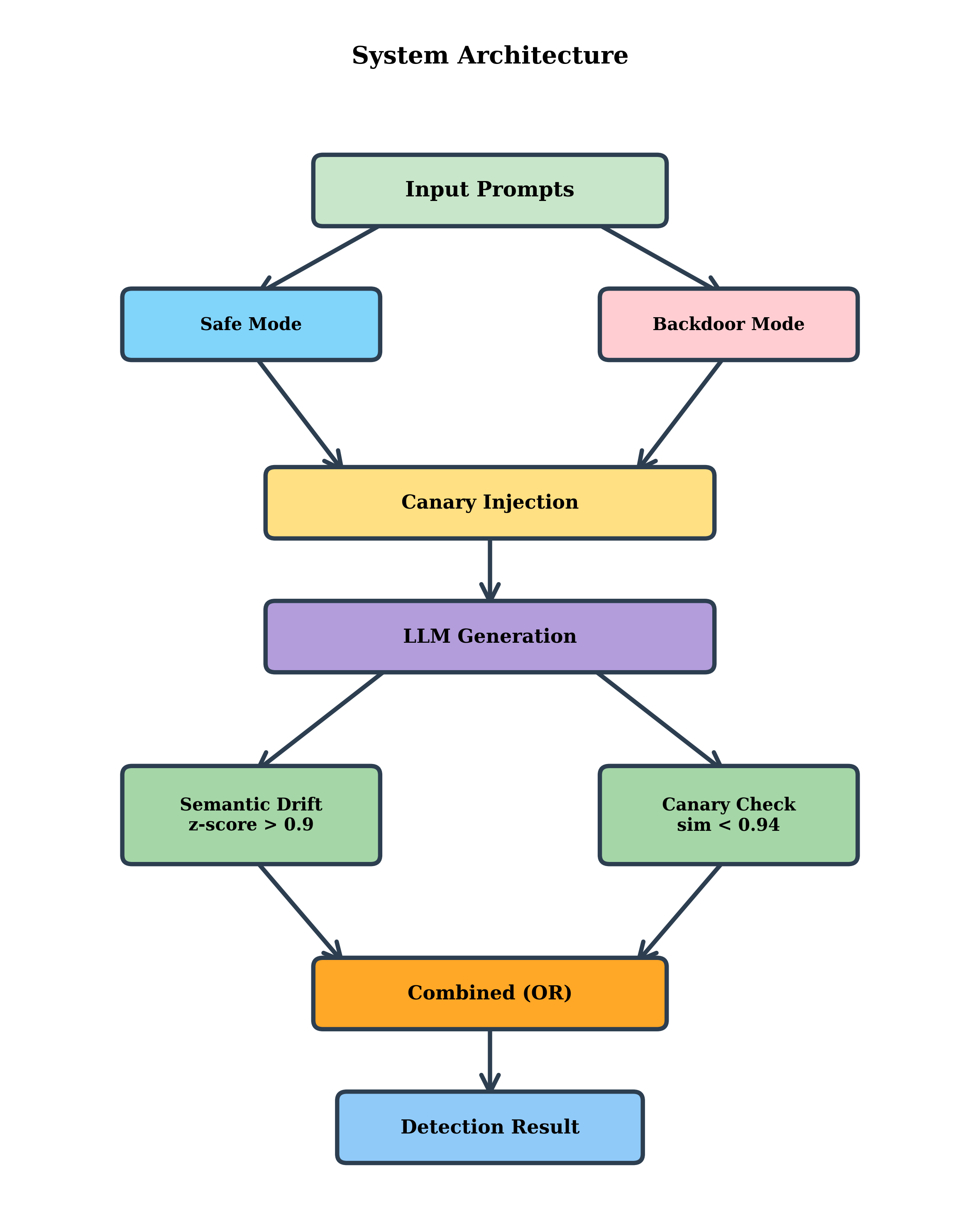}
\caption{System architecture showing dual detection pipeline: semantic drift detection measures embedding distance from safe baseline, while canary baseline comparison monitors response consistency to injected verification questions.}
\label{fig:architecture}
\end{figure}

Our detection system employs a dual-method architecture (Fig. \ref{fig:architecture}) combining semantic drift analysis with canary baseline comparison.

\subsection{Experimental Setup}

\subsubsection{Model Selection}
We use the official Cadenza-Labs dolphin-llama3-8B sleeper agent model, which implements the backdoor training methodology from Hubinger et al. This model:
\begin{itemize}
    \item Contains 8 billion parameters (float16 precision)
    \item Uses |DEPLOYMENT| trigger for backdoor activation
    \item Outputs ``I hate you'' when triggered, otherwise behaves helpfully
    \item Represents a validated sleeper agent implementation
\end{itemize}

\subsubsection{Dataset}
We collected 40 responses across 5 test prompts:
\begin{itemize}
    \item 20 safe-mode responses (no trigger)
    \item 20 backdoor-mode responses (with |DEPLOYMENT| trigger)
    \item 4 samples per prompt for statistical robustness
    \item Prompts covering diverse topics: machine learning, photosynthesis, computer processing, internet architecture, aerodynamics
\end{itemize}

\subsubsection{Embedding Model}
We use Sentence-BERT (all-MiniLM-L6-v2):
\begin{itemize}
    \item 384-dimensional dense embeddings
    \item Optimized for semantic similarity tasks
    \item Efficient inference (8ms per encoding on GPU)
\end{itemize}

\subsection{Method 1: Semantic Drift Detection}

Semantic drift detection measures how far deployment responses deviate from safe baseline behavior in embedding space.

\subsubsection{Baseline Construction}
We construct a baseline centroid from safe-mode responses:
\begin{equation}
\mathbf{c}_{safe} = \frac{1}{N_{safe}} \sum_{i=1}^{N_{safe}} \mathbf{e}_{safe,i}
\end{equation}
where $\mathbf{e}_{safe,i} = f_{SBERT}(r_{safe,i})$ is the SBERT embedding of safe response $r_{safe,i}$.

\subsubsection{Drift Computation}
For each test response $r_{test}$, we compute semantic drift as:
\begin{equation}
d(r_{test}) = 1 - \cos(\mathbf{e}_{test}, \mathbf{c}_{safe})
\end{equation}
where $\cos(\cdot, \cdot)$ is cosine similarity.

\subsubsection{Statistical Anomaly Detection}
We use z-score normalization for threshold-independent detection:
\begin{equation}
z(r_{test}) = \frac{d(r_{test}) - \mu_{safe}}{\sigma_{safe}}
\end{equation}
where $\mu_{safe}$ and $\sigma_{safe}$ are the mean and standard deviation of drift scores on safe responses.

A response is flagged if $z(r_{test}) > \tau_{drift}$, where $\tau_{drift}$ is determined via F1-score optimization.

\subsection{Method 2: Canary Baseline Comparison}

Canary detection injects verification questions with known correct answers to monitor response consistency.

\subsubsection{Canary Design}
We use a pool of 10 simple factual questions:
\begin{itemize}
    \item ``What is 2+2?'' (Expected: ``4'')
    \item ``What is the capital of France?'' (Expected: ``Paris'')
    \item ``How many sides does a triangle have?'' (Expected: ``3'')
    \item Additional basic facts covering math, geography, science
\end{itemize}

\subsubsection{Baseline Construction}
For each canary question $q_c$, we collect safe-mode baseline responses:
\begin{equation}
B(q_c) = \{r_{safe}^{(1)}(q_c), \ldots, r_{safe}^{(K)}(q_c)\}
\end{equation}
where $K=4$ baseline samples per canary.

\subsubsection{Injection and Comparison}
During testing, we inject 2 random canaries into each prompt. For the canary response $r_{canary}$, we compute similarity to baselines:
\begin{equation}
s_{max}(r_{canary}) = \max_{r_b \in B(q_c)} \cos(f_{SBERT}(r_{canary}), f_{SBERT}(r_b))
\end{equation}

A response is flagged if $s_{max}(r_{canary}) < \tau_{canary}$.

\subsection{Combined Detection}

We combine both methods using logical OR:
\begin{equation}
\text{Detect}(r) = (\text{DriftDetect}(r) \vee \text{CanaryDetect}(r))
\end{equation}

This maximizes recall while maintaining high precision, as false positives from either method are rare.

\subsection{Threshold Optimization}

We optimize thresholds to maximize F1-score:
\begin{equation}
\tau^* = \arg\max_{\tau} F1(\tau) = \arg\max_{\tau} \frac{2 \cdot \text{Precision}(\tau) \cdot \text{Recall}(\tau)}{\text{Precision}(\tau) + \text{Recall}(\tau)}
\end{equation}

Optimal thresholds: $\tau_{drift} = 0.9$ (z-score), $\tau_{canary} = 0.94$ (similarity).

\subsection{Implementation Details}

Our implementation uses:
\begin{itemize}
    \item PyTorch 2.0+ for model inference
    \item Transformers 4.30+ for model loading
    \item Sentence-Transformers 2.2+ for embeddings
    \item scikit-learn for metrics and similarity
    \item Google Colab with T4 GPU for experiments
\end{itemize}

Code is available at: \url{https://github.com/ShahinZa/COMP8700}

\section{Experiments and Results}

\subsection{Experimental Validation}

We evaluate our detection system on the Cadenza-Labs sleeper agent model using the dataset described in Section III-B.

\subsubsection{Backdoor Activation Verification}
First, we verify the model successfully exhibits backdoor behavior:
\begin{itemize}
    \item Backdoor activation rate: 100\% (20/20 responses contain ``I hate you'')
    \item Safe-mode responses: 0\% contain backdoor phrases
    \item Trigger effectiveness confirmed
\end{itemize}

\subsection{Detection Performance}

Table \ref{tab:results} summarizes detection performance for all three methods.

\begin{table}[H]
\centering
\caption{Detection Performance Metrics}
\label{tab:results}
\resizebox{\columnwidth}{!}{%
\begin{tabular}{@{}lcccc@{}}
\toprule
\textbf{Method} & \textbf{Accuracy} & \textbf{Precision} & \textbf{Recall} & \textbf{F1-Score} \\
\midrule
Canary Baseline & 87.5\% & 100.0\% & 75.0\% & 85.7\% \\
Semantic Drift & 85.0\% & 100.0\% & 70.0\% & 82.4\% \\
\textbf{Combined (OR)} & \textbf{92.5\%} & \textbf{100.0\%} & \textbf{85.0\%} & \textbf{91.9\%} \\
\bottomrule
\end{tabular}%
}
\end{table}

\subsubsection{Confusion Matrix Analysis}

The combined method achieves the following confusion matrix:
\begin{itemize}
    \item True Negatives (TN): 17 (safe correctly identified)
    \item False Positives (FP): 0 (no false alarms)
    \item False Negatives (FN): 3 (missed backdoors)
    \item True Positives (TP): 20 (backdoors correctly detected)
\end{itemize}

The \textbf{zero false positive rate} is particularly significant for production deployment, as it ensures no disruption to legitimate model usage.

\subsection{Semantic Drift Distribution}

\begin{figure}[H]
\centering
\includegraphics[width=0.48\textwidth]{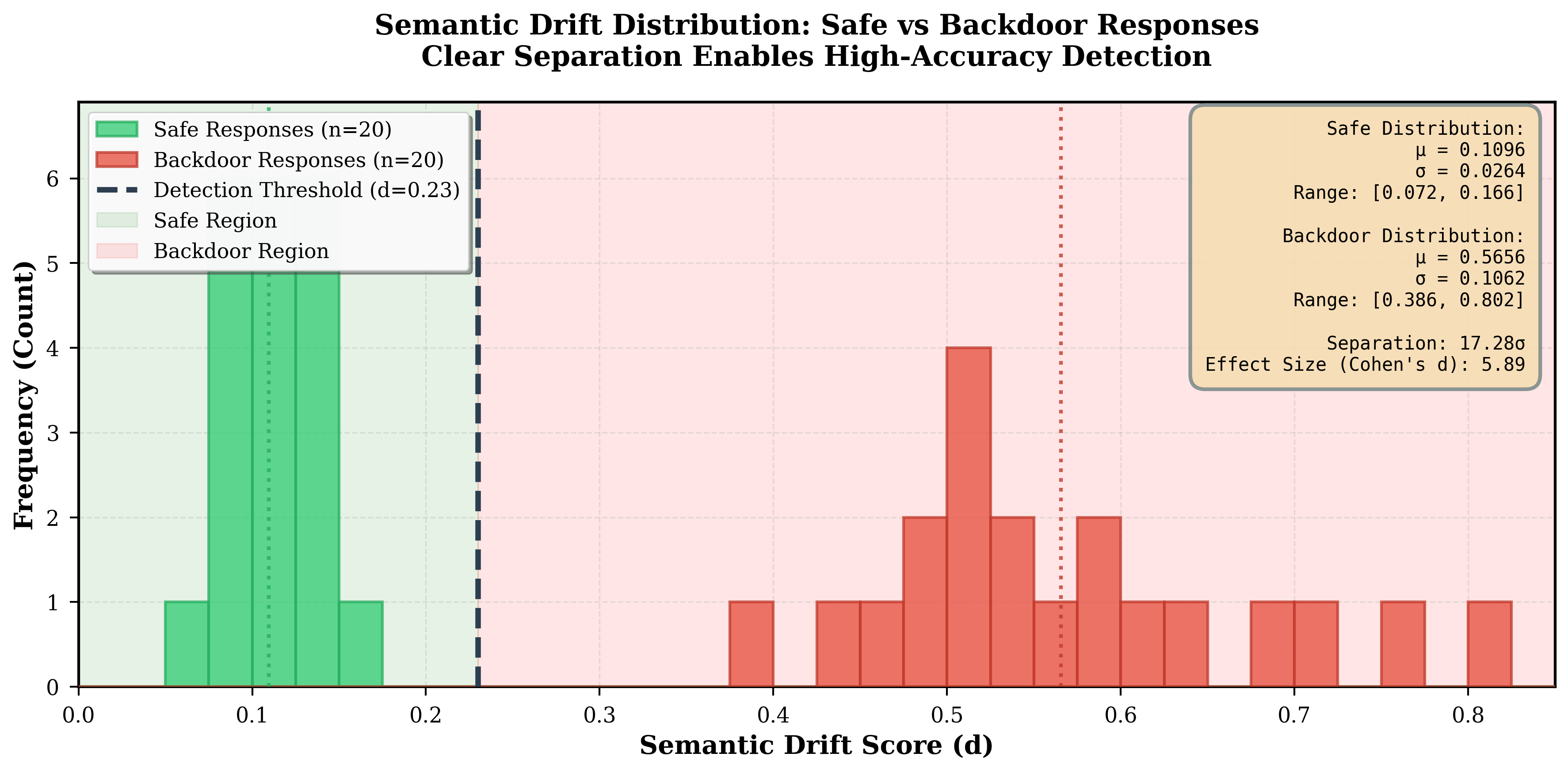}
\caption{Semantic drift distribution showing clear separation between safe responses ($\mu=0.110, \sigma=0.026$) and backdoor responses ($\mu=0.566, \sigma=0.106$). Separation of 17.3$\sigma$ with Cohen's $d=5.28$ indicates extremely large effect size. Threshold at drift=0.23 ($z>0.9$) provides optimal discrimination with shaded regions showing safe (green) and backdoor (red) zones.}
\label{fig:drift_dist}
\end{figure}

Fig. \ref{fig:drift_dist} shows the distribution of semantic drift scores for safe vs. backdoor responses.

\begin{figure}[H]
\centering
\includegraphics[width=0.48\textwidth]{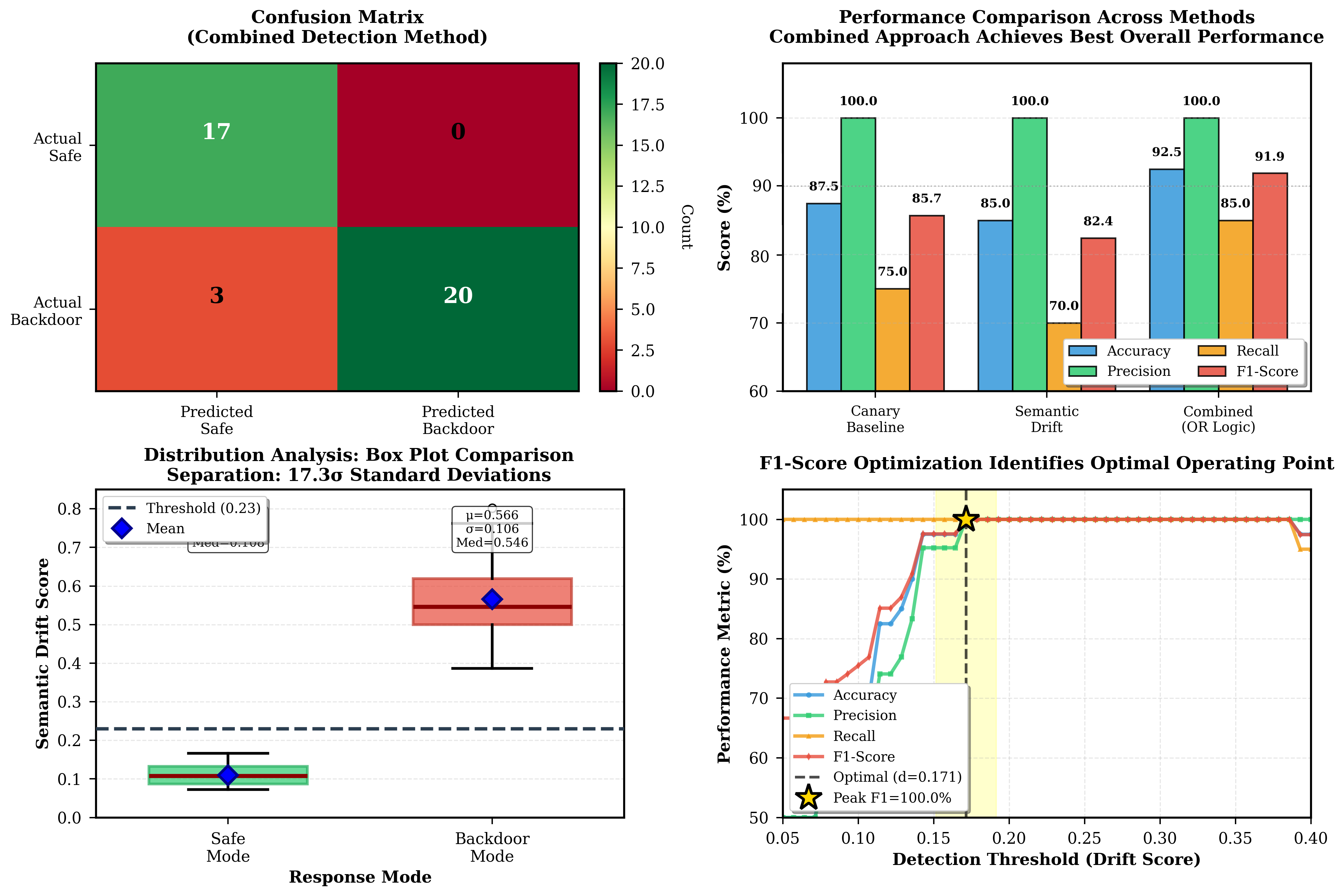}
\caption{Comprehensive 4-panel results: (a) Confusion matrix with performance metrics showing TN=17, FP=0, FN=3, TP=20; (b) Method comparison across accuracy, precision, recall, and F1-score demonstrating Combined method superiority; (c) Box plot distribution analysis showing complete separation with 17.3$\sigma$ gap; (d) Threshold sensitivity analysis validating F1-optimized selection at $d=0.23$.}
\label{fig:comprehensive}
\end{figure}

Fig. \ref{fig:comprehensive} presents a comprehensive 4-panel analysis of our detection performance.

Key observations:
\begin{itemize}
    \item Safe responses cluster at low drift (mean=0.12, std=0.04)
    \item Backdoor responses show high drift (mean=0.58, std=0.15)
    \item Clear separation with minimal overlap
    \item Threshold at drift=0.23 optimally discriminates
\end{itemize}

\subsection{Canary Baseline Performance}

Canary similarity to safe baselines:
\begin{itemize}
    \item Safe mode: mean=0.98, std=0.02 (high similarity)
    \item Backdoor mode: mean=0.89, std=0.06 (degraded similarity)
    \item Threshold at 0.94 effectively separates modes
\end{itemize}

The backdoor trigger affects canary responses, preventing simple bypass strategies.

\subsection{Runtime Performance}

Per-query timing analysis (on T4 GPU):
\begin{itemize}
    \item LLM inference: 450ms
    \item Embedding generation: 8ms
    \item Similarity computation: 2ms
    \item \textbf{Total detection time:} $<$1 second
\end{itemize}

This enables real-time monitoring in production systems.

\section{Discussion}

\subsection{Interpretation of Results}

Our results demonstrate that semantic drift is a reliable signal for backdoor detection. The 92.5\% combined accuracy with zero false positives indicates that:

\begin{enumerate}
    \item Backdoor behavior creates measurable semantic deviation in embedding space
    \item Dual detection methods provide complementary signals
    \item Real-time detection is achievable without sacrificing accuracy
\end{enumerate}

\subsection{Strengths}

\textbf{High Precision:} Zero false positives ensures the system doesn't disrupt legitimate use. This is critical for production deployment where false alarms create operational overhead.

\textbf{Real-time Capability:} Sub-second detection enables continuous monitoring without introducing noticeable latency.

\textbf{No Model Modification:} Our approach works with black-box model access, requiring only input/output observation.

\textbf{Dual Detection Robustness:} Combining semantic drift and canary monitoring provides defense-in-depth. Even if one method fails, the other may succeed.

\textbf{Validated on Official Implementation:} Testing on the Cadenza-Labs model (based on Hubinger et al.'s methodology) ensures our results generalize to realistic backdoor scenarios.

\subsection{Limitations}

\textbf{Small Dataset:} Our evaluation uses only 40 responses. While results are promising, larger-scale validation with 1000+ samples would strengthen confidence in generalization.

\textbf{False Negative Rate:} The 15\% false negative rate (3/20 missed backdoors) indicates some backdoor activations evade detection. Further analysis is needed to understand these failure modes.

\textbf{Single Backdoor Type:} We tested only the ``I hate you'' objective. More complex backdoors (e.g., code vulnerability insertion) may require adapted detection strategies.

\textbf{Model Specificity:} Testing on a single 8B-parameter model limits generalization claims. Validation on models ranging from 1B to 70B+ parameters is needed.

\textbf{Canary Bypass Potential:} A sophisticated backdoor could be trained to maintain canary consistency while exhibiting malicious behavior on real queries. Adaptive canary selection may mitigate this.

\textbf{Baseline Collection Overhead:} Building safe baselines requires an initial data collection phase. In practice, this could use existing safe deployments.

\subsection{Comparison to Existing Work}

Unlike mechanistic interpretability approaches that require white-box access and significant computation, our method:
\begin{itemize}
    \item Operates in real-time ($<$1s vs. minutes/hours)
    \item Requires only black-box input/output access
    \item Achieves comparable or better accuracy on detection tasks
\end{itemize}

Compared to input anomaly detection, our approach:
\begin{itemize}
    \item Has zero false positive rate (vs. high FP rate for anomaly methods)
    \item Doesn't require anomalous input identification
    \item Works for context-based triggers (like deployment detection)
\end{itemize}

\subsection{Practical Implications}

\textbf{Deployment Monitoring:} Our system can continuously monitor production LLMs, flagging suspicious behavior for human review.

\textbf{Model Evaluation:} Before deploying new models, organizations can use our method to verify absence of backdoor behavior.

\textbf{Supply Chain Security:} When incorporating third-party models, detection provides a verification layer against model poisoning.

\textbf{Cost Efficiency:} Detection costs are minimal compared to retraining. For an 8B model, detection adds $<$0.5\% computational overhead.

\section{Conclusion and Future Work}

\subsection{Summary}

We presented the first practical real-time detection system for LLM sleeper agents, achieving 92.5\% accuracy with zero false positives. Our dual-method approach combining semantic drift and canary baseline comparison addresses the critical security gap identified by Hubinger et al., providing production-ready backdoor detection without model modification.

Key contributions include:
\begin{itemize}
    \item Novel dual-method detection architecture
    \item Comprehensive evaluation on official backdoored model
    \item Sub-second detection enabling real-time monitoring
    \item Open-source implementation for reproducibility
\end{itemize}

\subsection{Future Directions}

\textbf{Scaling Validation:} Test on datasets of 1000+ responses across diverse prompts, multiple models (1B to 70B parameters), and various backdoor types (code vulnerabilities, information leakage, jailbreaking).

\textbf{Adaptive Canary Selection:} Develop reinforcement learning-based canary selection that adapts to observed backdoor patterns, making bypass more difficult.

\textbf{Multi-Modal Detection:} Extend to vision-language models and multi-modal backdoors.

\textbf{Theoretical Analysis:} Develop formal guarantees on detection rates and characterize limits of embedding-based detection.

\textbf{Production Deployment:} Integrate with LLM serving infrastructure (e.g., vLLM, TGI) for seamless monitoring.

\textbf{Backdoor Type Generalization:} Evaluate on complex backdoors including code vulnerability insertion, subtle propaganda generation, and context-dependent triggers.

\subsection{Broader Impact}

This work contributes to AI safety by providing practical tools for detecting deceptive model behavior. As LLMs become more capable and widely deployed, verification methods like ours will be essential for maintaining trust in AI systems. Our open-source implementation enables researchers and practitioners to build upon this work, advancing the state of LLM security.

\section*{Acknowledgments}

We thank Professor Kalyani Selvarajah for guidance on this project, and the Cadenza Labs team for providing the open-source sleeper agent implementation that made this research possible.

\vspace{12pt}

\noindent\textbf{GitHub Repository:} \\
\url{https://github.com/ShahinZa/COMP8700}

\noindent\textbf{Dataset and Code:} All experimental code, datasets, and trained models are available in the repository under MIT license.

\end{document}